%% file: main.tex
\documentclass[conference]{IEEEtran}
\IEEEoverridecommandlockouts
\usepackage{array}
\usepackage{cite}
\usepackage{censor}

\StopCensoring

\usepackage{amsmath,amssymb,amsfonts}
\usepackage{algorithmic}
\usepackage{booktabs}
\usepackage{graphicx}
\usepackage{multirow, array}
\usepackage{subcaption}
\usepackage{makecell}
\usepackage{url}
\usepackage{hyperref}

\def\BibTeX{{\rm B\kern-.05em{\sc i\kern-.025em b}\kern-.08em
    T\kern-.1667em\lower.7ex\hbox{E}\kern-.125emX}}
\begin{document}

\title{Learning Adaptive Force Control for Contact-Rich Sample Scraping with Heterogeneous Materials
\\
% \thanks{This work was supported by the Royal Academy of Engineering under the Research Fellowship Scheme and Google DeepMind Research Ready Programme.}
}

% \author{\IEEEauthorblockN{\censor{Cenk Cetin}}
% \IEEEauthorblockA{
% \and
% \IEEEauthorblockA{
% \and
% \IEEEauthorblockA{

% \author{}
\author{\censor{Cenk Cetin$^{1}$, Shreyas Pouli$^{1}$, Gabriella Pizzuto$^{1, 2}$}\\
\thanks{\censor{$^{1}$ School of Computer Science \& Informatics, University of Liverpool, UK.}}%
\thanks{\censor{$^{2}$ Department of Chemistry, University of Liverpool, UK.}} %
\thanks{\censor{This work was supported by the Royal Academy of Engineering under the Research Fellowship Scheme and Google DeepMind Research Ready programme.}}
}

% \and
% \and
% \and

\maketitle

\begin{abstract}

The increasing demand for accelerated scientific discovery, driven by global challenges, highlights the need for advanced AI-driven robots.
Robotic systems in this area are still in their infancy, often being highly specialised and hand-engineered. 
This limits their application in early-stage materials discovery, where sample properties are heterogeneous and unpredictable. 
Deploying robotic chemists in human-centric labs is key for the next horizon of autonomous discovery, as complex tasks still demand the dexterity of human scientists. 
Robotic manipulation in this context is uniquely challenged by handling diverse materials (from granular to cohesive), under varying lab conditions. 
For example, humans effortlessly use spatulas for scraping materials from vial walls.
Automating this process is challenging because it goes beyond simple robotic insertion tasks and traditional lab automation, requiring the execution of fine-grained movements within a constrained environment (the sample vial).
Our work proposes an adaptive control framework to address this, relying on a low-level Cartesian impedance controller for stable and compliant physical interaction and a high-level reinforcement learning agent that learns to dynamically adjust interaction forces at the end-effector. 
The agent is guided by perception feedback, which provides the material's location.
We first created a task-representative simulation environment with a Franka Research 3 robot, a scraping tool, and a sample vial containing heterogeneous materials. 
To facilitate the learning of an adaptive policy and model diverse characteristics, the material is modelled as a collection of spheres, where each sphere is assigned a unique dislodgement force threshold, which is procedurally generated using Perlin noise. 
We train an agent to autonomously learn and adapt the optimal contact wrench for a sample scraping task in simulation and then successfully transfer this policy to a real robotic setup. 
Our method was evaluated across five different material setups, outperforming a fixed-wrench baseline by an average of 10.9\%.

% We show how our method performs across 5 different material setups and demonstrate how it outperforms the baseline with fixed wrench profiles by an average of 10.9\%. 
\end{abstract}

\begin{figure}[]
    \centering
    \includegraphics[width=0.49\textwidth]{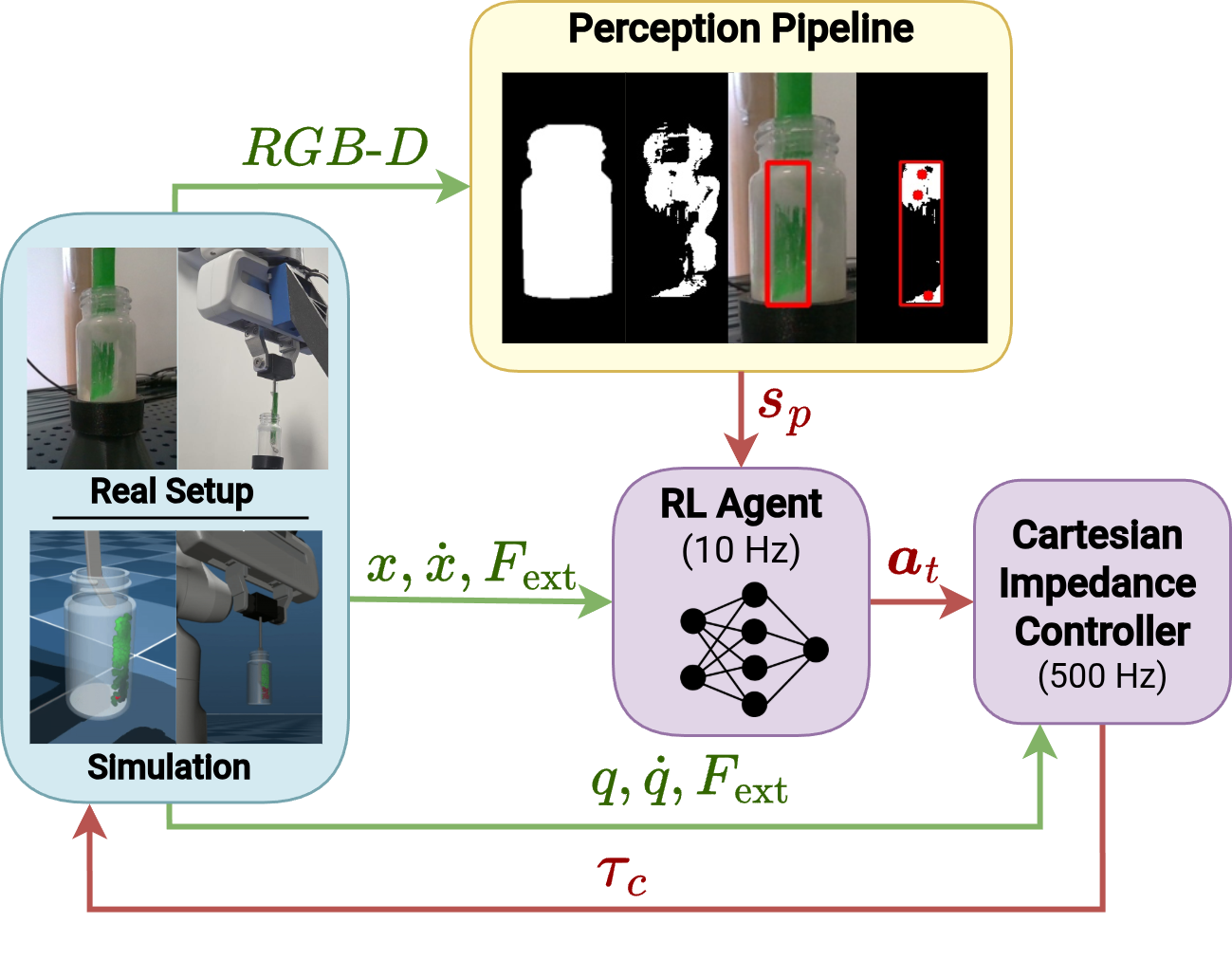}
    \caption{The proposed control architecture for autonomous sample scraping. A high-level RL policy receives a visual state ($s_{\text{p}}=[v_1,v_2,v_3]^T$), with $v_i=[c_{ix},c_{iy},c_{iz},p_i]^T$ denoting each cluster's centroid and residue percentage) and the robot's Cartesian state with the external wrench ($x, \dot{x}, F_{\text{ext}}$). The policy outputs a hybrid action command ($\boldsymbol{a}_t = [f_x^c, \tau_y^c, z^D]^T$) at 10~Hz, defining a desired force, torque, and position in the z-axis. This command is used as a goal to the Cartesian impedance controller running at 500~Hz, which generates the compliant joint torques ($\tau_c$).}

    \label{fig:pipeline}
\end{figure}
% \begin{IEEEkeywords}
% component, formatting, style, styling, insert
% \end{IEEEkeywords}

\input{sections/introduction}
\input{sections/related_work}
\input{sections/methodology}

\input{sections/experimental_evaluation}

\input{sections/conclusions}

\bibliographystyle{IEEEtran}
\bibliography{bibliography}

\end{document}

%% file: sections/introduction.tex
\section{Introduction}
\label{section:introduction}

Transforming materials discovery is crucial for tackling topical issues~\cite{Stein2019}, from clean energy to pharmaceuticals. 
Traditionally, this process has been slow and costly due to its heavy reliance on manual, repetitive, and often hazardous experimental workflows performed by human chemists. 
The subjective nature of these manual experiments has historically led to inconsistencies, contributing to a reproducibility crisis in the field~\cite{Tom2024}. 
To accelerate fundamental breakthroughs, there is a growing paradigm shift towards using intelligent autonomous systems, with a diverse array of general-purpose robots being deployed to automate workflows like synthesis~\cite{Dai2024} and photocatalysis~\cite{Burger2020}. 
While this has improved efficiency, an open gap remains: current systems are often limited to pre-programmed sample transportation. 
This hinders fully autonomous, end-to-end operation in scientific labs, where robots need to adapt to novel protocols and handle heterogeneous materials, which is a common challenge in early-stage discovery. 
Efficient and reliable handling of materials, from fine powders to granular crystals, is essential for chemistry lab automation. 

% [Current works to date and open problem] - What other people have done and why this is not good enough
To date, advanced robot chemists primarily use position controllers~\cite{Pizzuto2024Scraping, Darvish2025} even for contact-rich manipulation tasks.
These controllers are ill-suited for tasks requiring compliance and adaptation to variable forces. 
Impedance controllers~\cite{hoganImpedanceControlApproach1984}, in contrast, provide the necessary force control and adaptability to manage unpredictable interactions, making them ideal for handling heterogeneous materials. 
% These controllers are ill-suited for tasks requiring compliance and adaptation to variable forces. 
These controllers are especially valuable for tasks like sample scraping~\cite{Pizzuto2024Scraping} in automated solid-state materials chemistry~\cite{Lunt2024} and drug development. 
This process is essential for recovering expensive molecules and requires a precise vertical motion to retrieve powdered or crystalline samples from vial walls. 
The task demands adaptive force control to handle variable material properties \textit{e.g.}, adhesive and cohesive, as well as visual feedback to gauge how much material remains. 
The challenge is further compounded by the tools themselves. 
Standard laboratory spatulas are often compliant and bendable, which complicates relying solely on reactive force control. 
The force measured at the robot's wrist does not directly correspond to the force at the tool tip due to tool deformation. 
Prior works on automating this process~\cite{Pizzuto2024Scraping} have not addressed these requirements.

% [How you address this open gap and covering what you did in brief]
We address these challenges with a novel adaptive control framework that combines a low-level Cartesian Impedance Controller (CIC) with a high-level Reinforcement Learning (RL) agent (Fig.~\ref{fig:pipeline}). 
The RL agent learns a perception-based, feedforward wrench command that is robust to both unmodelled tool and material dynamics.
The RL agent learns an optimal force policy in real-time by interacting with the environment, allowing it to adapt to varying material properties like hardness and adhesion. 
To ensure the robot applies force only where it is needed, we integrate visual (RGB-D) feedback, which informs the agent of the sample's location.
By isolating force learning from high-dimensional joint control, our method simplifies the learning process and enhances sim-to-real transfer. 
This framework paves the way for developing adaptive, force-aware robotic chemists for laboratory automation, enabling previously infeasible operations such as scraping heterogeneous material samples.

In summary, the contributions of this work are:

(1) A novel adaptive control framework for combining a low-level Cartesian impedance controller with a high-level RL agent for learning the desired interaction wrenches to perform material removal from vial walls during autonomous sample scraping. This work presents a perception-driven, `force-aware’ approach to the challenging problem of in-vial manipulation with heterogeneous materials;

% (2) The development of an RL agent for learning adaptive force and torque applications. This policy effectively bypasses the need for explicit analytical models of material properties;

(2) A multi-stage perception pipeline to autonomously localise the vial and accurately detect the sample material for the scraping task. This enables the robot to execute the task without a priori knowing the material's position or distribution;

(3) A demonstration and empirical evaluation of the learned robot behaviour both in simulation and in a real-world chemistry lab across different materials.

%% file: sections/related_work.tex
\section{Related Work}
\label{sec:related_work}

\subsection{Robot Skill Learning for Laboratory Automation}
\label{ssec:skill_learning_related_work}

Self-driving laboratories are evolving from simple open-loop automation towards adaptive, modular, and reconfigurable robotic systems~\cite{Tom2024}. 
While perception-guided long-horizon laboratory skills have been proposed for some tasks~\cite{Darvish2025}, these often rely on basic manipulation skills that primarily involve the pick-and-place of laboratory objects, without handling materials.
While these systems are adept at performing tedious and repetitive tasks, they lack the dexterity and skill of human chemists. 
Current approaches for acquiring these laboratory skills (\textit{i.e.}, low-level behaviours) typically fall into two categories: either they are learned within physics-based digital twins using reinforcement learning (RL) and then transferred to the real setup~\cite{Pizzuto2024Scraping, Radulov2025}, or they use multi-modal sensory data with pre-defined motions for handling materials~\cite{Nakajima2023} and glassware~\cite{Butterworth2023, Fakhruldeen2025}.
Our work builds on previous research in laboratory skill acquisition for handling heterogeneous materials (using an RL approach similar to~\cite{Pizzuto2024Scraping, kadokawa2023, Radulov2025}). 
However, we specifically address the challenge of materials adhering to the inside of vials. 
To overcome this, we propose an RL agent that adapts external forces, which is a critical factor for success in these tasks. 
Furthermore, unlike prior work, our method uses multi-modal perception to inform the agent of the material's location.

\subsection{Learning-Based Adaptive Control}
\label{ssec:adaptive_control}

Many learning-based adaptive control methods, particularly those using RL, have focused on adapting control parameters (\textit{e.g.}, stiffness, damping, and desired forces)~\cite{martin2019, Yang2022}. 
While these approaches use robot-environment interaction data to improve performance, tuning these parameters often leads to large action spaces, sample inefficiency, and poor sim-to-real transfer, especially when handling difficult-to-model dynamics. 
Prior works also do not explicitly train the RL agent to learn a target Cartesian wrench ($\boldsymbol{F_c^{ext}}$), which is critical for tasks requiring real-time adjustment to material resistance and adhesion.
Our work addresses this by specifically having the RL agent learn a policy that outputs the target Cartesian wrench ($\boldsymbol{F_c^{ext}}$). 
This learned wrench is then compliantly executed by an underlying Cartesian impedance controller with fixed or optimised parameters. 
This separation from low-level parameter tuning simplifies the RL action space, allowing the agent to focus on the interaction dynamics needed to adapt the applied wrench to different material properties.

% For instance, previous frameworks~\cite{martin2019} have used algorithms like Proximal Policy Optimisation (PPO)~\cite{schulman2017} to learn complex motion and impedance parameters.

% While RL has been extensively explored for adaptive robot control, its application to contact-rich tasks like sample scraping, which requires adaptive wrench control for varying material properties, presents unique challenges. 
% Many existing RL methods either learn direct joint commands, which struggle with the complexities of contact and compliance, or they learn high-dimensional controller parameters. 

%% file: sections/methodology.tex
\section{Methodology}
\label{sec:methodology}
This work introduces a novel RL framework with visual feedback for the challenging, contact-rich task of in-vial sample scraping with heterogeneous materials.
This task is routinely performed in chemistry labs and is fundamental for experimental workflows such as powder diffraction~\cite{Lunt2024}.

% Our proposed method combines a low-level Cartesian impedance controller with a high-level RL agent for  learning the desired interaction wrenches using visual perception to inform it about the location of the materials.

% As illustrated in , the robotic manipulator is equipped with a scraping tool, grasped by its parallel gripper, and a vial is positioned within its workspace.
% The robot's task is to learn how to insert the tool, scrape the vial's interior walls, and remove the material inside.

% Force feedback is particularly intriguing as monitoring fine powder manipulation through perception is challenging due to the ubiquitous
% nature of the task: transparent media e.g. lab glassware suffer
% from specular reflections, and adding white, or transparent
% crystals makes it even more problematic. 
% In addition, given
% the contact-rich nature of the task, we believe this can
% only be captured through force feedback rather than visual
% demonstrations or feedback as different materials exhibit
% different properties when exposed to environmental factors.

%\begin{figure}[h!]
%    \centering
%    \includegraphics[width=0.24\textwidth]{figures/arm_vial.png}
%    \caption{Representation of the scraping task, showing the tool-vial interaction geometry during in-vial material removal.}
%    \label{fig:scraping_representation}
%\end{figure}

\begin{figure}[h!]
    \centering
    \includegraphics[width=0.48\textwidth]{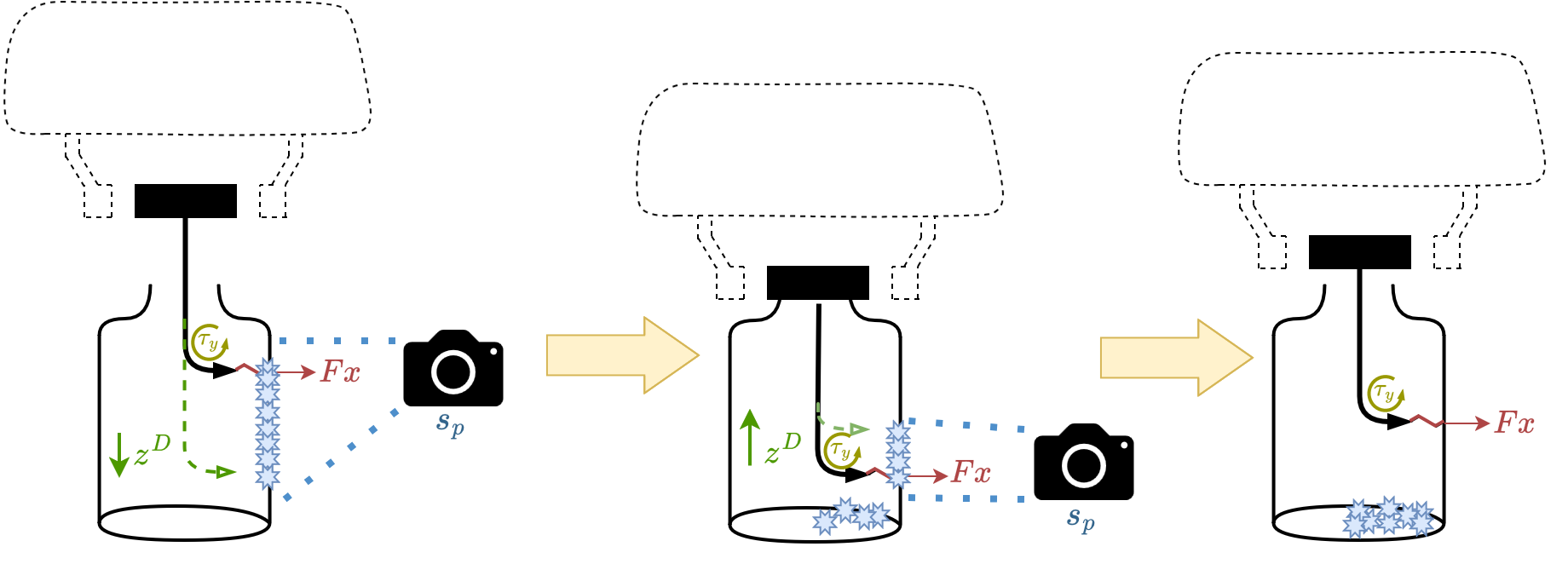}
    \caption{The sequential phases of the autonomous in-vial sample scraping task: (\textit{left}) tool insertion with lateral force $F_x$ and torque $\tau_y$ applied against the vial wall, with RGB-D camera capturing visual state $s_p$; (\textit{centre}) RL policy adapts $F_x$, $\tau_y$, and vertical position $z^{D}$ based on real-time perception feedback $s_p$, controlling both contact wrench and scraping direction; (\textit{right}) final upward sweep with adjusted $F_x$ to clear residual material, with dislodged particles settled at the bottom of the vial.}
    \label{fig:scraping_representation}
\end{figure}

\subsection{Task Formulation}
\label{ssec:task_formulation}

The task \emph{scrape}, as defined in previous works~\cite{Pizzuto2024Scraping}, involves the robot reaching the bottom of a vial while maintaining contact with the wall to remove material (Fig.~\ref{fig:scraping_representation}). 
This constitutes a fundamental manipulation subtask within a broader laboratory workflow; once dislodged from the vial walls, gravity causes the material to settle at the bottom, where it can be subsequently extracted or processed.
% Figure~\ref{fig:scraping_representation} illustrates the scraping setup and contact geometry considered in this work.
The end-effector commands and state readings are computed with respect to the tool centre point (TCP) at the tip of the scraping tool.
The task is deemed successful if the robot removes all material from the target window, confirmed by visual feedback.

\subsection{Cartesian Impedance Controller}
\label{ssec:CIC}
To ensure safe and compliant interaction during the contact-rich scraping task, we employ a Cartesian impedance controller~\cite{Hogan1985}. 
This control strategy is essential for tasks involving brittle glassware as it regulates the robot's end-effector to exhibit a desired mass-spring-damper behaviour in response to external forces.
The controller operates on the robot's rigid-body dynamics, expressed in the joint space $\boldsymbol{q} \in \mathbb{R}^{n}$:

\begin{equation} \label{eq:robot_dynamics}
\boldsymbol{M}(\boldsymbol{q})\ddot{\boldsymbol{q}} + \boldsymbol{C}(\boldsymbol{q}, \dot{\boldsymbol{q}})\dot{\boldsymbol{q}} = \boldsymbol{\tau}_{c} + \boldsymbol{\tau}_{ext}
\end{equation}

In this formulation, $\boldsymbol{M(q)}$ is the inertia matrix, $\boldsymbol{C(q, \dot{q})}$ captures the Coriolis and centripetal forces, $\boldsymbol{\tau}_{ext}$ represents external torques, and $\boldsymbol{\tau}_{c}$ is the commanded torque; gravity torques are omitted because they are compensated by the robot's internal low-level controllers.
The commanded torque $\boldsymbol{\tau}_{c}$ is generated from the control law in Equation~(\ref{eq:torque_composition}), which superimposes three distinct torque components\cite{mayr2024cpp}:

\begin{equation} \label{eq:torque_composition}
    \boldsymbol{\tau}_{c} = \boldsymbol{\tau}_{c}^{ca} + \boldsymbol{\tau}_{c}^{ns} + \boldsymbol{\tau}_{c}^{ext}
\end{equation}

The Cartesian impedance torque ($\boldsymbol{\tau}_{c}^{ca}$) provides the core compliant behaviour~\cite{hoganImpedanceControlApproach1984}. 
The nullspace torque ($\boldsymbol{\tau}_{c}^{ns}$) allows the manipulator to manage its joint configuration to avoid joint limits while the end-effector remains in contact with the vial wall, which is essential for a continuous $360^{\circ}$ scraping motion. 
The commanded external wrench torque ($\boldsymbol{\tau}_{c}^{ext}$) enables the robot to apply a desired force and torque to its environment. 
This is calculated using the equation:
\begin{equation} \label{eq:wrench_torque}
    \boldsymbol{\tau}_{c}^{ext} = \boldsymbol{J}(\boldsymbol{q})^{T}\boldsymbol{F}_{c}^{ext}
\end{equation}
where $\boldsymbol{J}(\boldsymbol{q})$ is the robot's Jacobian and $\boldsymbol{F}_{c}^{ext}$ is the desired Cartesian wrench in the end-effector frame. 
% This combined approach allows the robot to perform the scraping task with precise contact, adaptable pose, and controlled force application.

Our proposed methodology builds on this such that an RL agent learns the commanded Cartesian wrench, $\boldsymbol{F}_{c}^{ext}$, whilst keeping the underlying impedance parameters of the CIC fixed to provide a stable, low-level controller.
% This approach simplifies the learning problem by tasking the RL agent to learn the high-level interaction wrench, rather than low-level controller gains, which aids in sim-to-real transfer.

\begin{figure*}[h!]
    \centering
    \includegraphics[width=0.9\textwidth]{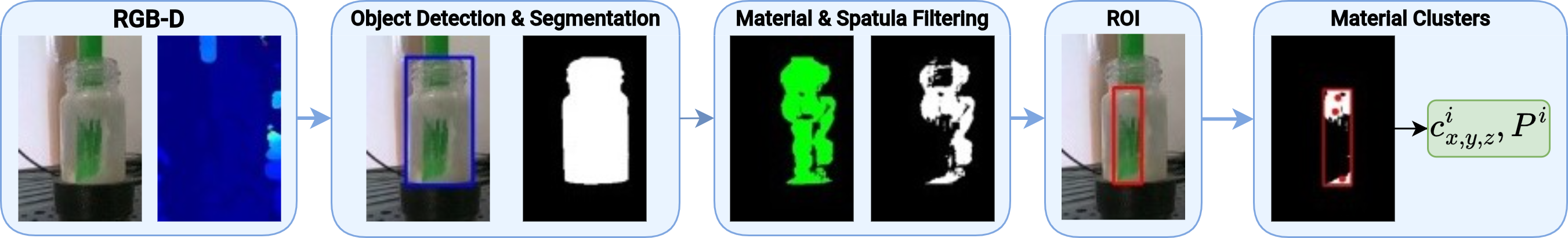}
    \caption{The proposed perception pipeline. From RGB-D input, YOLO~\cite{Redmon2016} localises the vial, segmented via GrabCut~\cite{Chen2008}. Depth filtering isolates front-facing material and colour-based filtering removes the spatula. K-means clustering over a region of interest (ROI) then yields the material clusters, each represented by a centroid ($c_x, c_y, c_z$) and residue percentage ($p$).}

    \label{fig:perception_pipeline}
\end{figure*}

\subsection{Reinforcement Learning for Cartesian Wrench Generation}
\label{ssec:problem_formulation}

For modelling the underlying decision-making problem of an agent learning $\boldsymbol{F}_{c}^{ext}$, we adopt the finite-horizon discounted Markov decision process (MDP) formalism~\cite{Bellman1957},~\cite{Puterman1994}. 
It can be defined as a tuple $\mathcal{M} = (\mathcal{S}, \mathcal{A}, \mathcal{P}, r, \gamma, \mathcal{T})$, consisting of a continuous state space $S$, a continuous action space $A$, a transition function $P(s_{t+1}|s_t, a_t)$ defining the probability of the environment transitioning to state $s_{t+1}$ after the agent takes action $a_t$ in state $s_t$, a reward function $r(s_t,a_t)$ specifying the scalar reward the agent receives after taking action $a_t$ in state $s_t$, a discount factor $\gamma \in [0, 1)$, and a horizon $H$ specifying the number of steps per episode.
The actions are drawn from a stochastic policy which is a state-conditioned probability distribution over actions $\pi(a_t|s_t)$.
The goal of the agent is to find the optimal policy $\pi^*$, which maximises the expected sum of discounted future rewards $ \mathbb{E}[\sum_{i=0}^T \gamma^{i} r_{t+i}]$ when the agent selects actions in each state according to $\pi^*$.

\textbf{State Space} ($\mathcal{S}$).
The state $s_t$ is a vector providing key observations for the scraping task, comprising three main components.
First, the end-effector state includes the Cartesian position and orientation of the tool tip, along with the scalar magnitude of its velocity. 
Second, the estimated external wrench provides direct feedback on the interaction with the environment through the forces ($F_x, F_y, F_z$) and torques ($T_x, T_y, T_z$).
% derived from the robot's dynamics model.
Finally, the sphere clusters provide information about the remaining sample, represented by the 3D centroids and residue percentages for three distinct clusters of material, as provided by the perception pipeline (Fig.~\ref{fig:perception_pipeline}) on the real system and by the simulated materials (Fig.~\ref{fig:sim_env_spheres}). 

\textbf{Action Space} ($\mathcal{A}$). The action $\boldsymbol{a}_t \in \mathbb{R}^3$ selected by the RL agent is a hybrid command that specifies the desired forces, torques, and positions along different Cartesian axes of the end-effector. 
This approach simplifies the learning task by focusing on the most critical degrees of freedom for scraping.
Specifically, the action vector is defined as $\boldsymbol{a}_t = [f_x^c, \tau_y^c, z^D]^T$. The first component, $f_x^c$, is the commanded force along the x-axis, used to maintain appropriate contact and apply a consistent normal force against the vial wall. 
The second component, $\tau_y^c$, is the commanded torque around the y-axis, which generates the tangential scraping motion required to dislodge material. 
Finally, $z^D$ is the desired position along the z-axis, controlling the vertical sweeping movement of the tool inside the vial. 
This decoupled, hybrid action space allows the agent to learn the distinct skills of maintaining surface (vial wall) contact, scraping, and traversing more effectively.

\textbf{Reward Function ($r$)}.
The reward combines an efficiency objective with sparse task-completion and safety shaping terms. The total reward $r_t$ at each timestep $t$ is given by:
% \begin{equation} \label{eq:reward_function}
% \begin{split}
%     r_t = \frac{\Delta m_t}{\|\boldsymbol{F}_{ext,t}\|_2 + \epsilon} + \lambda_E R_E - \lambda_C R_C
% \end{split}
% \end{equation}

\begin{equation} \label{eq:reward_function}
\begin{split}
    r_t = R_M + R_E - \lambda_C R_C
\end{split}
\end{equation}

Here, $R_M = \frac{\Delta m_t}{\|\boldsymbol{F}_{ext,t}\|_2}$ where $\Delta m_t$ is the newly removed material at timestep $t$, $\boldsymbol{F}_{ext,t}$ is the measured interaction wrench magnitude.
The milestone term $R_E$ is a sparse bonus granted at predefined milestones to prevent the agent from converging on overly conservative policies that exploit the simulation dynamics (\textit{e.g.}, moving slowly to minimise force penalties without progressing material removal), by rewarding task advancement.
The dense efficiency-based reward $R_M$ serves as the primary learning signal, and the magnitude of $R_E$ was determined empirically.
The contact penalty term $R_C$ penalises unintended collisions between non-functional tool parts (\textit{e.g.}, the spatula handle) and the vial, weighted by $\lambda_C = 0.01$, promoting safer interaction.
These terms motivate efficient material removal with minimum force towards task completion and collision-aware behaviour.
%The primary component is the progress reward $R_{P}$, which is proportional to the amount of newly scraped material and directly encourages task completion. 
% A velocity reward $R_{V}$ promotes the continuous vertical motion necessary for scraping; this reward is gated by contact with the material (\textit{i.e.}, with the inner wall). 
% For efficiency, $R_{T}$ is given for earlier faster progress, with its magnitude decaying over time.  
%A milestone reward $R_{E}$ is given upon reaching 90\% completion to encourage thorough cleaning. 
% This counteracts the agent's tendency to prematurely stop scraping by reducing its applied force to avoid the excessive force penalty $R_{F}$. 
% The milestone reward incentivises the agent to apply the necessary force to remove the final, more difficult material.
%Conversely, penalties are introduced to prevent undesirable behaviour; for example, a body contact penalty ($R_{C}$) for collisions between the vial and non-functional parts of the tool, such as the spatula handle.
% n excessive force penalty ($R_{F}$) is applied for interaction forces exceeding a predefined threshold and 

Ablation studies established the necessity of all terms in our reward function for achieving optimal performance with respect to completing the task effectively, safely, and efficiently. 
The body contact penalty weight $\lambda_C$ was the sole hyperparameter formally optimised via Bayesian optimisation to balance task performance against safe contact behaviour.

\subsection{Perception for Monitoring Task Progression}
\label{ssec:perception}

Our autonomous sample-scraping robotic system uses a multi-stage perception pipeline (Fig.~\ref{fig:perception_pipeline}) to localise a sample vial and detect the material's location. 

\begin{figure}[h]
    \centering
    \includegraphics[width=0.34\textwidth]{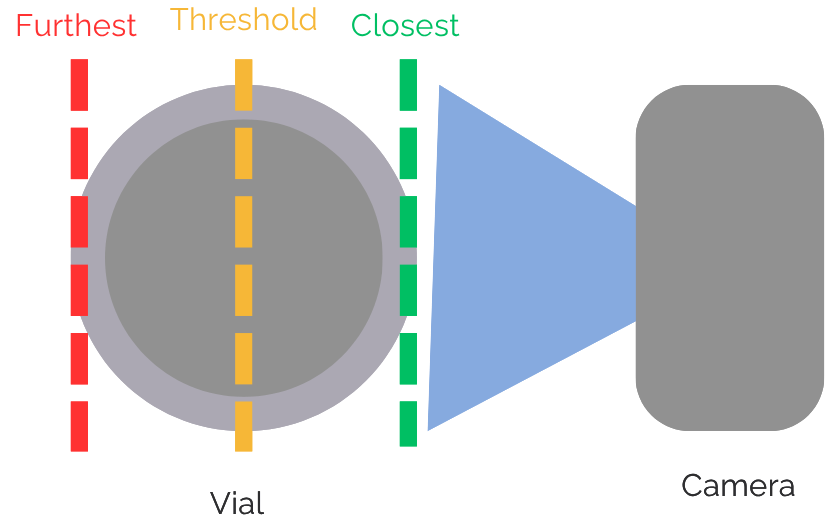}
    \caption{A dynamic depth threshold method to detect the presence of material closest to the camera. The threshold is calculated by interpolating between the closest and furthest depth measurements based on a user-defined ratio (\textit{e.g.}, a ratio of 0.5 would correspond to the front half of the vial.)}
    \label{fig:depth_thresholding}
\end{figure}

This pipeline was designed to provide online perception feedback for the robot controller, prioritising computational efficiency over granular accuracy. 
The approach begins with the acquisition of a multimodal data stream from an RGB-D camera mounted on the robot's end-effector. 
This is followed by an object detection and localisation task, where the glassware, typically a sample vial, is localised using an efficient deep-learning model like the you only look once (YOLO) architecture~\cite{Redmon2016}. 
The output of this stage is a 2D bounding box that tightly encloses the vial in the RGB-D image plane.
Subsequently, fine-grained image segmentation of the vial from its background is performed, assuming the material is on the vessel. 
Using the bounding box output, we apply the GrabCut algorithm~\cite{Chen2008}, a graph-cut-based method that models the foreground (vial) and background using a Gaussian mixture model. 
This detector-plus-graph-cut combination was chosen as a computationally efficient alternative to heavier end-to-end segmentation models.
The resulting binary mask is used to capture complementary depth information, precisely isolating the depth data corresponding to the vial region, including the vial, and the material on its front and back walls. 
For our task, the robot needs to accurately identify and scrape the material's front-facing surface, eliminating the risk of targeting the occluded back region, which is an ill-posed target for the robot manipulator. 
To isolate the front-facing scraping surface and the material on it, we employ a depth-thresholding method, illustrated in Fig.~\ref{fig:depth_thresholding}.

Our pipeline corrects for noise by analysing the mean ($\mu_z$) and standard deviation ($\sigma_z$) of the depth values, removing outliers defined as any depth value $z$ such that:
\begin{equation}
    \label{eqn:noise_correction}
    z \notin [\mu_z - \sigma_z, \mu_z + \sigma_z]
\end{equation}

This is followed by converting the depth mask into a binary contour map, which is applied to the original colour image to define the material regions. 
However, this region may also contain the spatula, as it is often indistinguishable from the material based on depth alone. 
Hence, to simplify the perception pipeline, we use colour information to separate the tool from the material. 
Pixels in the GrabCut mask are first clustered in RGB space via k-means, and each pixel is assigned to its nearest centroid. The centroid colours are then converted to HSV and thresholded by hue and saturation to remove the spatula, whose colour is intentionally distinct from the materials. A second k-means clustering stage is then applied to the remaining material pixels to obtain a compact, fixed-size representation for policy input.

This filtered mask is then used for region of interest (ROI) cropping, which corresponds to a percentage of the cropped image from the bounding box. 
Using a percentage-based crop instead of a pixel-based one allows for consistent cropping regardless of the vial's size or distance from the camera. 
This also provides a smaller window, improving visibility of the scraping area and simplifying the contact task from a curved surface to a plane. 
For this second clustering stage, centroids are computed from all white pixels in the filtered mask. The number of pixels assigned to each centroid is then used to compute cluster-wise percentage coverage over the ROI. 
Finally, each cluster centroid ($c_x, c_y$) is obtained from the RGB pixel coordinates, with $c_z$ retrieved from the corresponding depth pixel; these are transformed to world coordinates using the robot's kinematic chain and stored alongside their percentage coverage ($p$) as structured input for the RL policy.  The end-to-end perception pipeline is illustrated in Fig.~\ref{fig:perception_pipeline}.

%% file: sections/experimental_evaluation.tex
\section{Experimental Evaluation}
\label{sec:experiments}

In this section, we present a thorough evaluation of our proposed methodology across a wide range of materials, with our experiments designed to address the following research questions: (1) Can a Cartesian impedance controller with fixed wrench profiles be effectively used for contact-rich tasks in chemical lab automation, specifically for sample scraping with varying materials? (2) Can an RL policy be trained to autonomously learn and adapt the optimal contact wrench for a sample scraping task, enabling it to generalise across various materials with previously unknown force profiles? (3) How does the integration of visual and depth feedback enable localisation of materials inside sample vials? (4) How does the overall system perform when deployed in a real-world chemistry laboratory environment?

\begin{table*}[h!]
    \caption{Qualitative results for the five material samples, from the original sample to those obtained after scraping by the baseline (fixed wrench) and RL (adaptive wrench) methods.}
    \centering
    \begin{tabular}{l *{5}{c}}
        \toprule
        & \textbf{Liquid Dough} & \textbf{Liquid Cornflour} & \textbf{Dried Cornflour} & \textbf{Crystalline Salt} & \textbf{Crystalline Sugar} \\
        \midrule
        Original Sample & \includegraphics[width=0.20\textwidth, height=2.5cm, keepaspectratio]{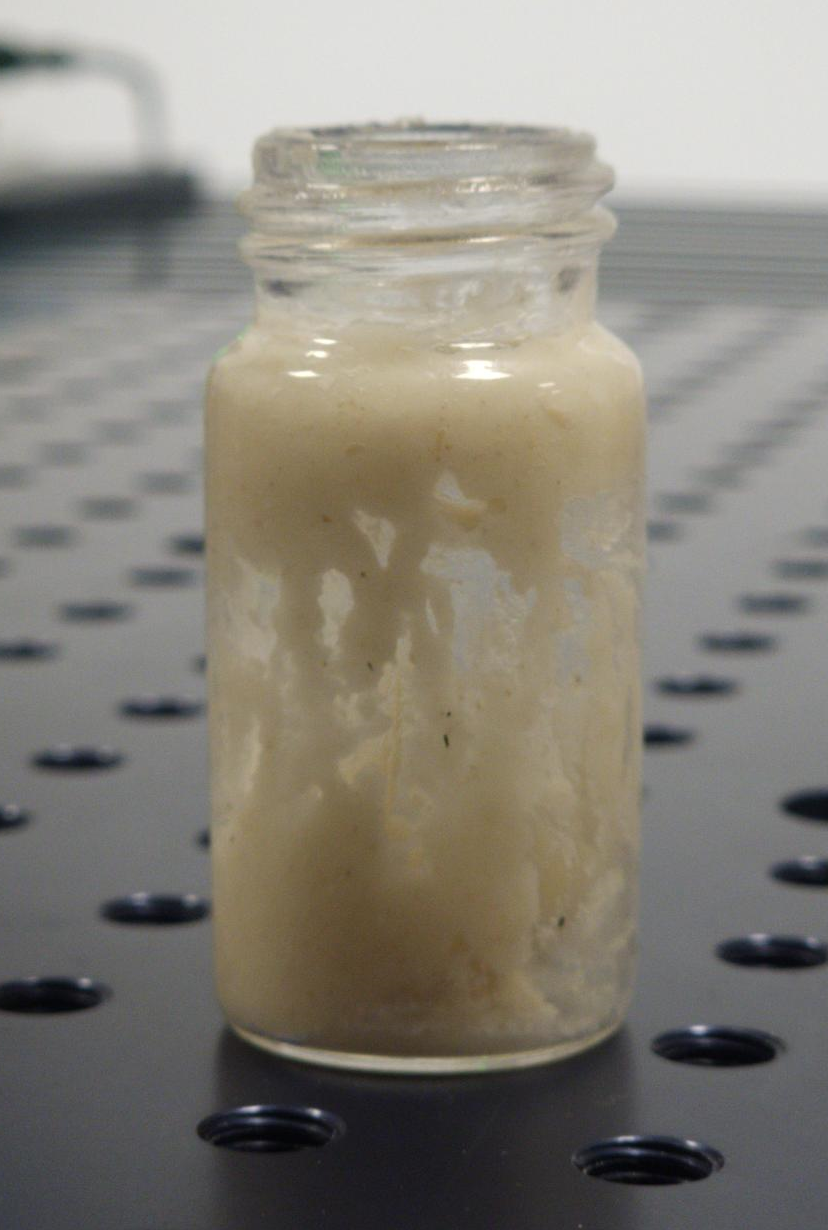} & \includegraphics[width=0.20\textwidth, height=2.5cm, keepaspectratio]{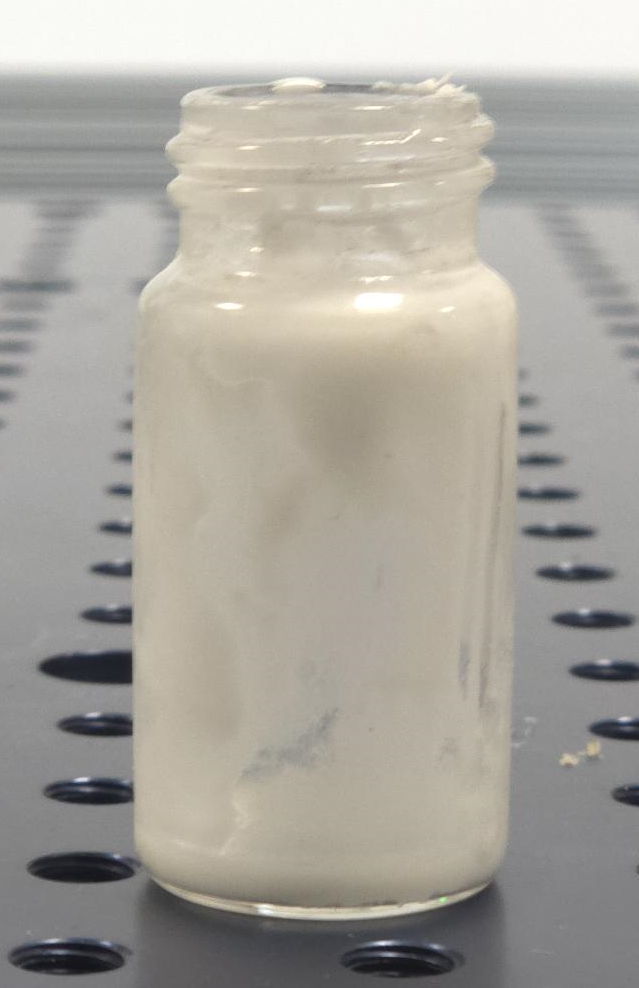} & \includegraphics[width=0.20\textwidth, height=2.5cm, keepaspectratio]{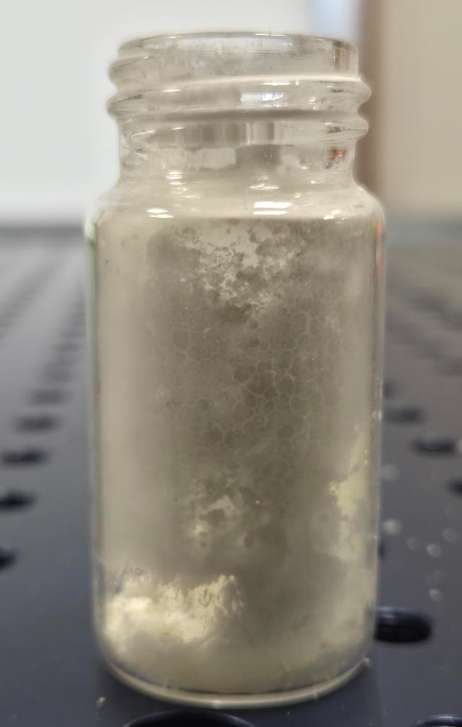} & \includegraphics[width=0.20\textwidth, height=2.5cm, keepaspectratio]{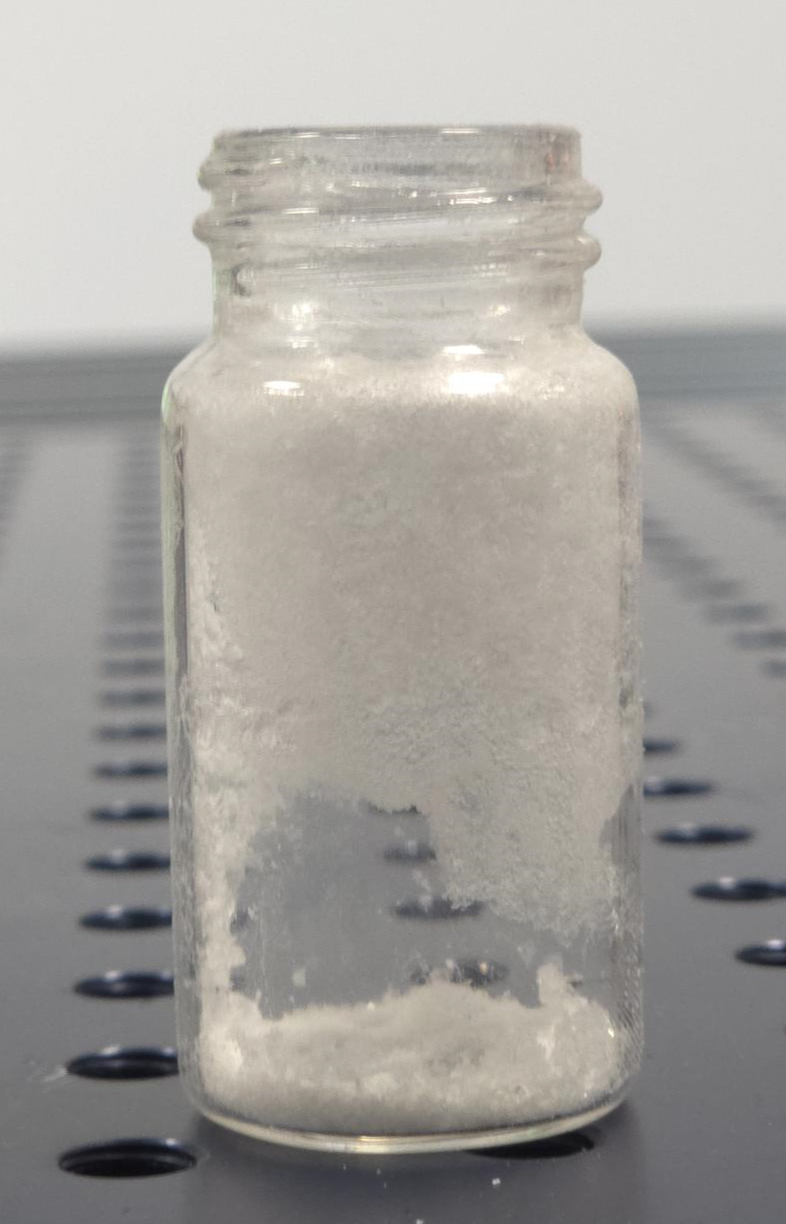} & \includegraphics[width=0.20\textwidth, height=2.5cm, keepaspectratio]{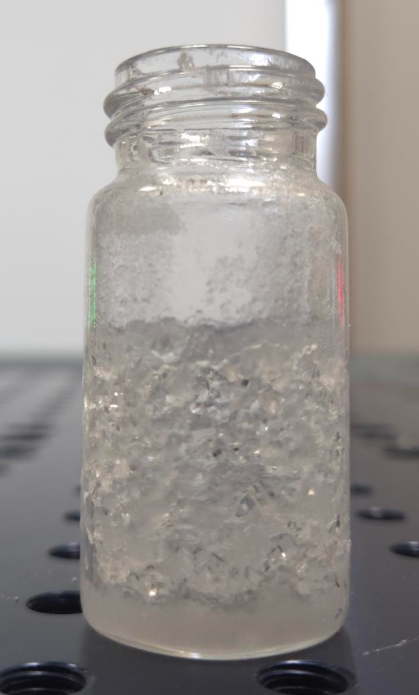} \\
        
        Robotic Scraped Sample (Baseline) & \includegraphics[width=0.20\textwidth, height=2.5cm, keepaspectratio]{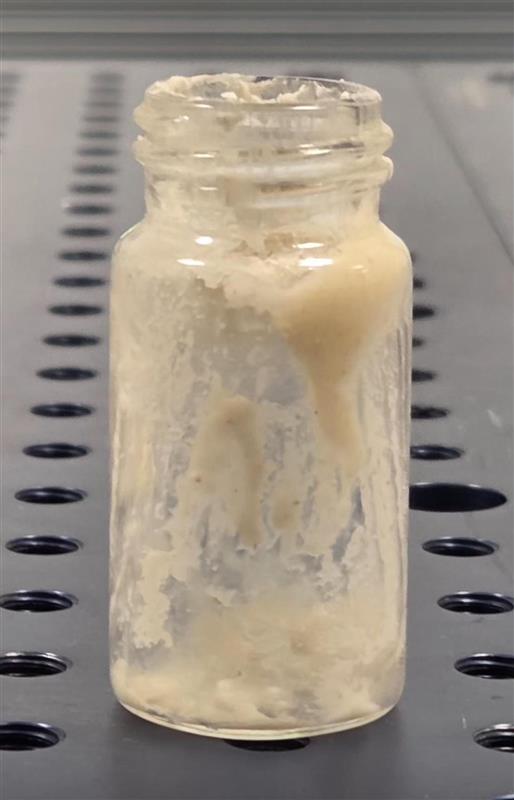} & \includegraphics[width=0.20\textwidth, height=2.5cm, keepaspectratio]{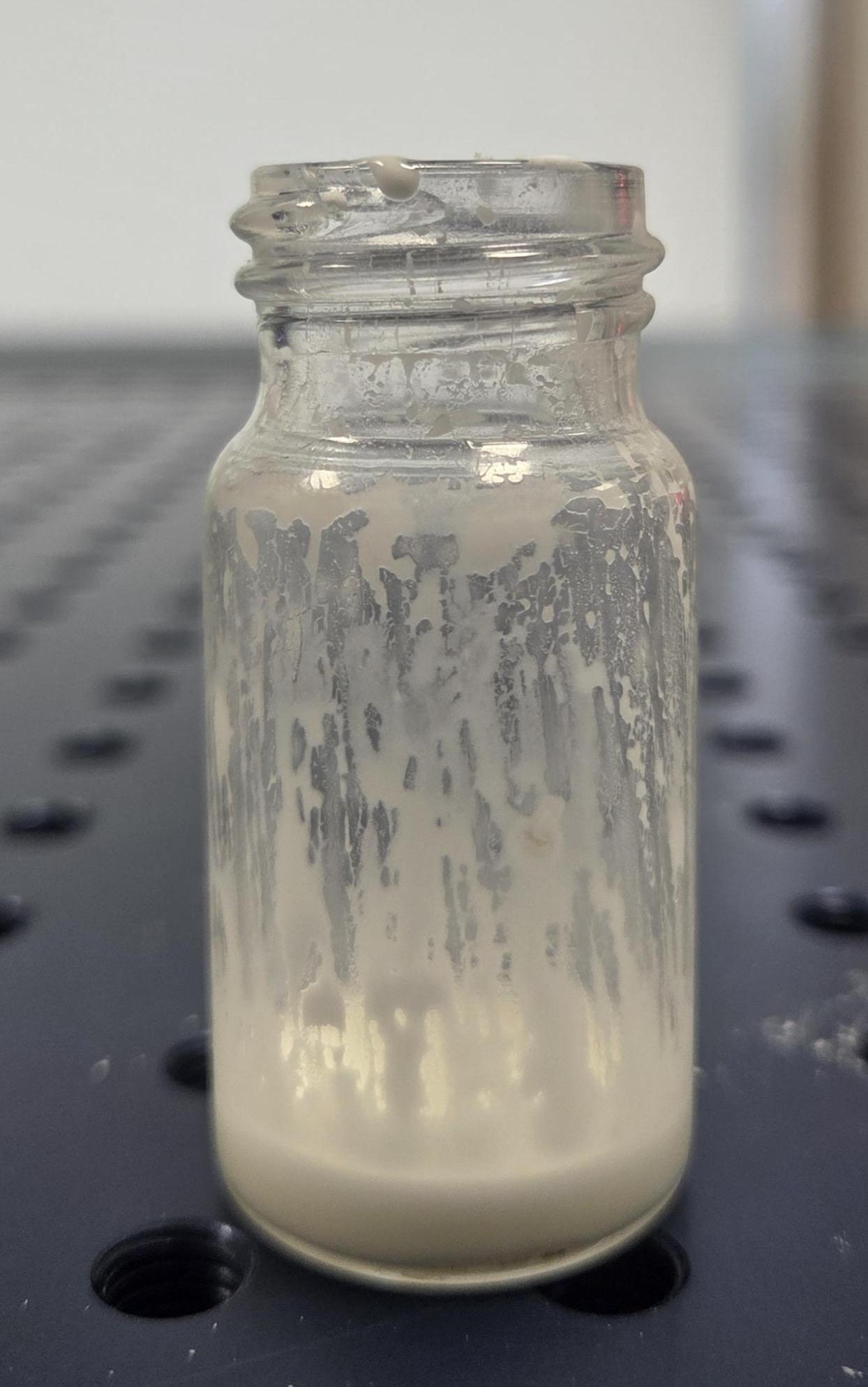} & \includegraphics[width=0.20\textwidth, height=2.5cm, keepaspectratio]{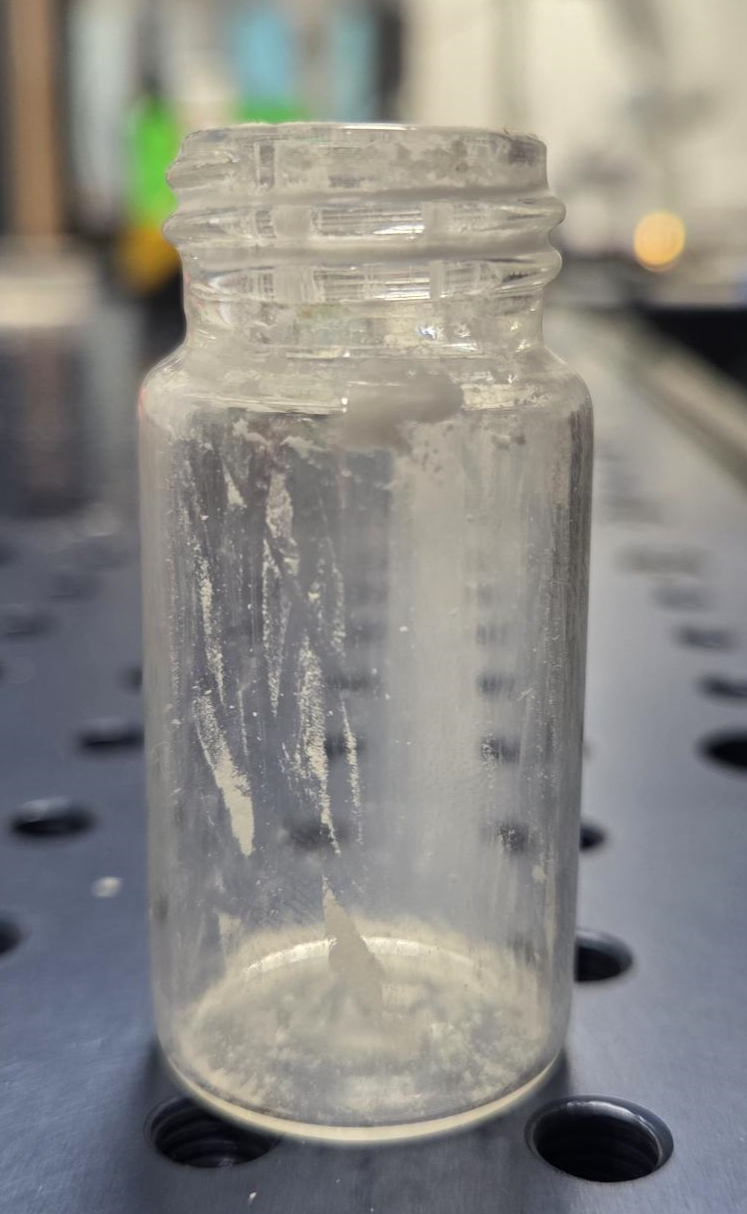} & \includegraphics[width=0.20\textwidth, height=2.5cm, keepaspectratio]{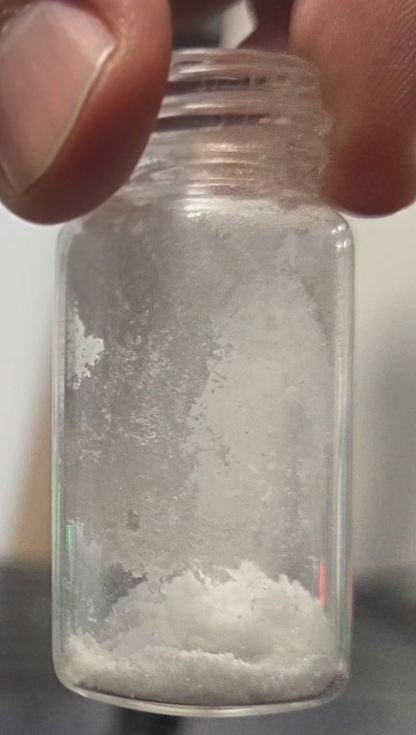} & \includegraphics[width=0.20\textwidth, height=2.5cm, keepaspectratio]{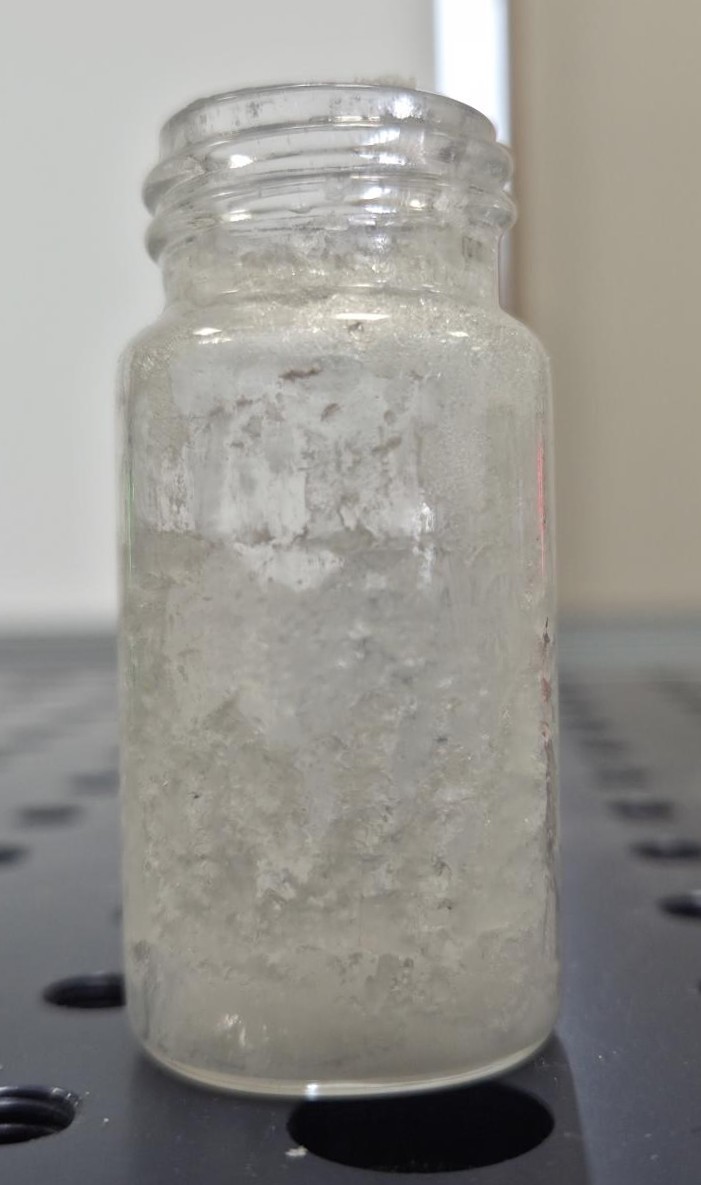} \\
        
        Robotic Scraped Sample (RL Method) & \includegraphics[width=0.20\textwidth, height=2.5cm, keepaspectratio]{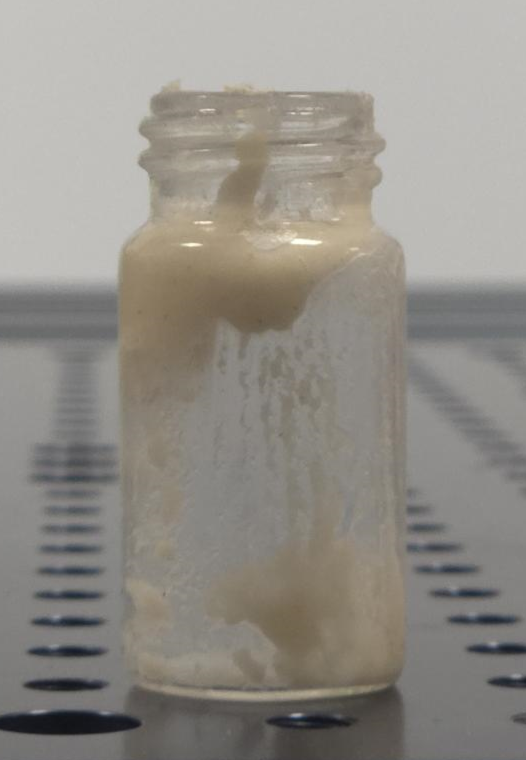} & \includegraphics[width=0.20\textwidth, height=2.5cm, keepaspectratio]{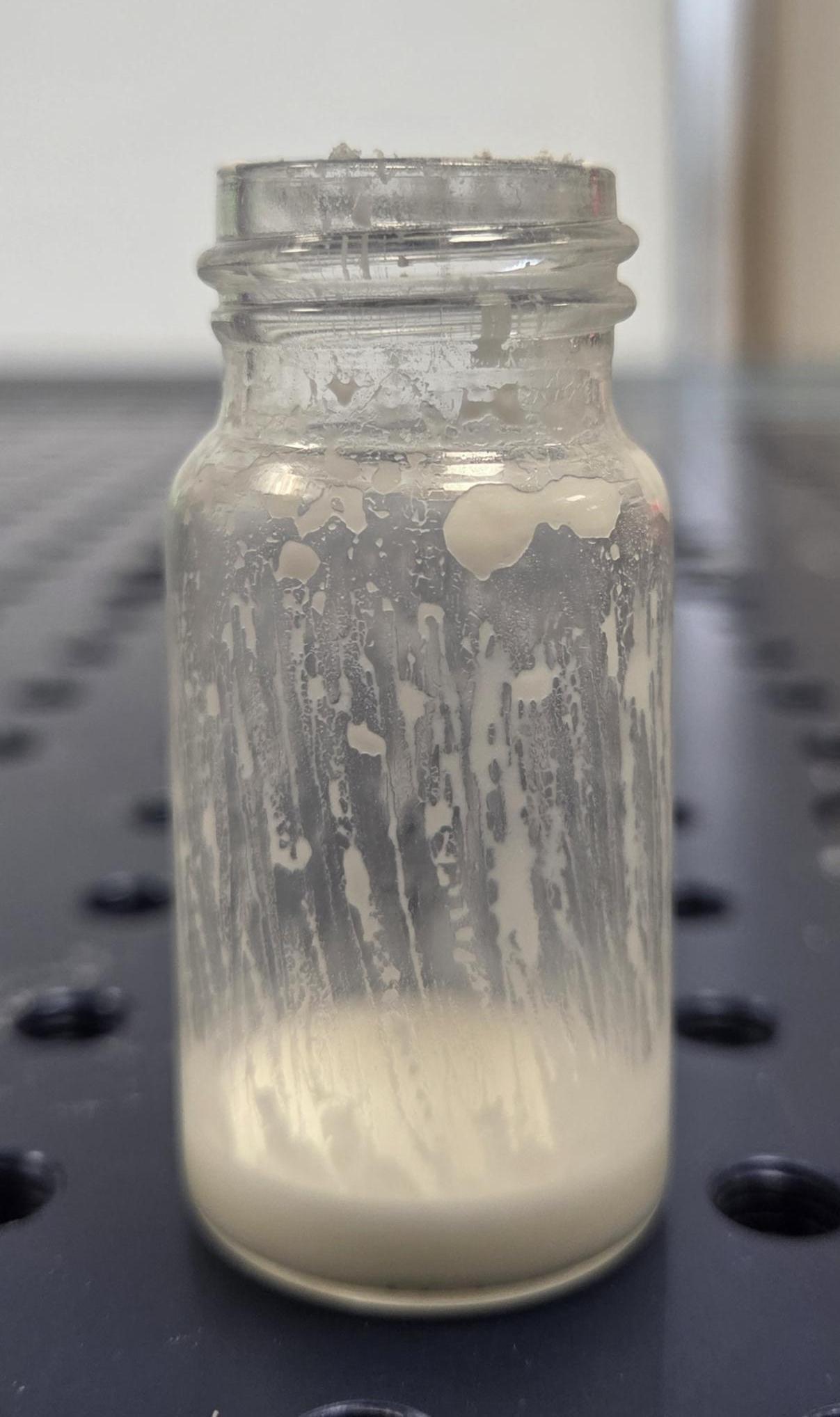} & \includegraphics[width=0.20\textwidth, height=2.5cm, keepaspectratio]{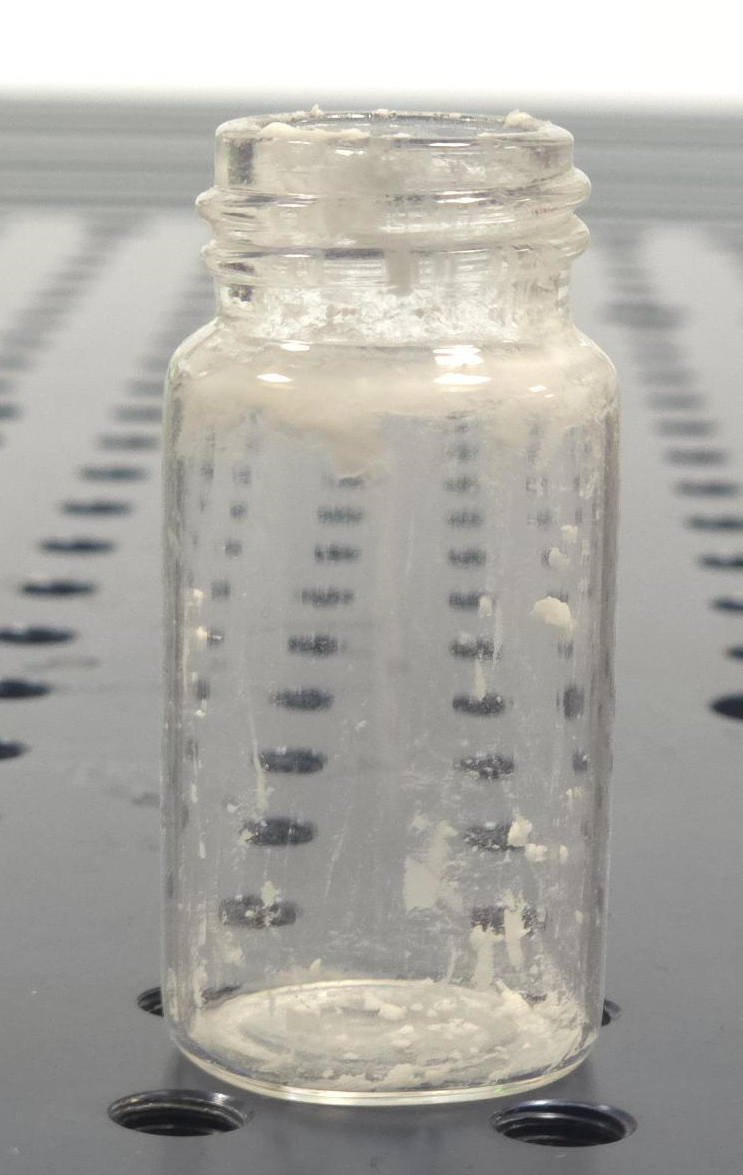} & \includegraphics[width=0.20\textwidth, height=2.5cm, keepaspectratio]{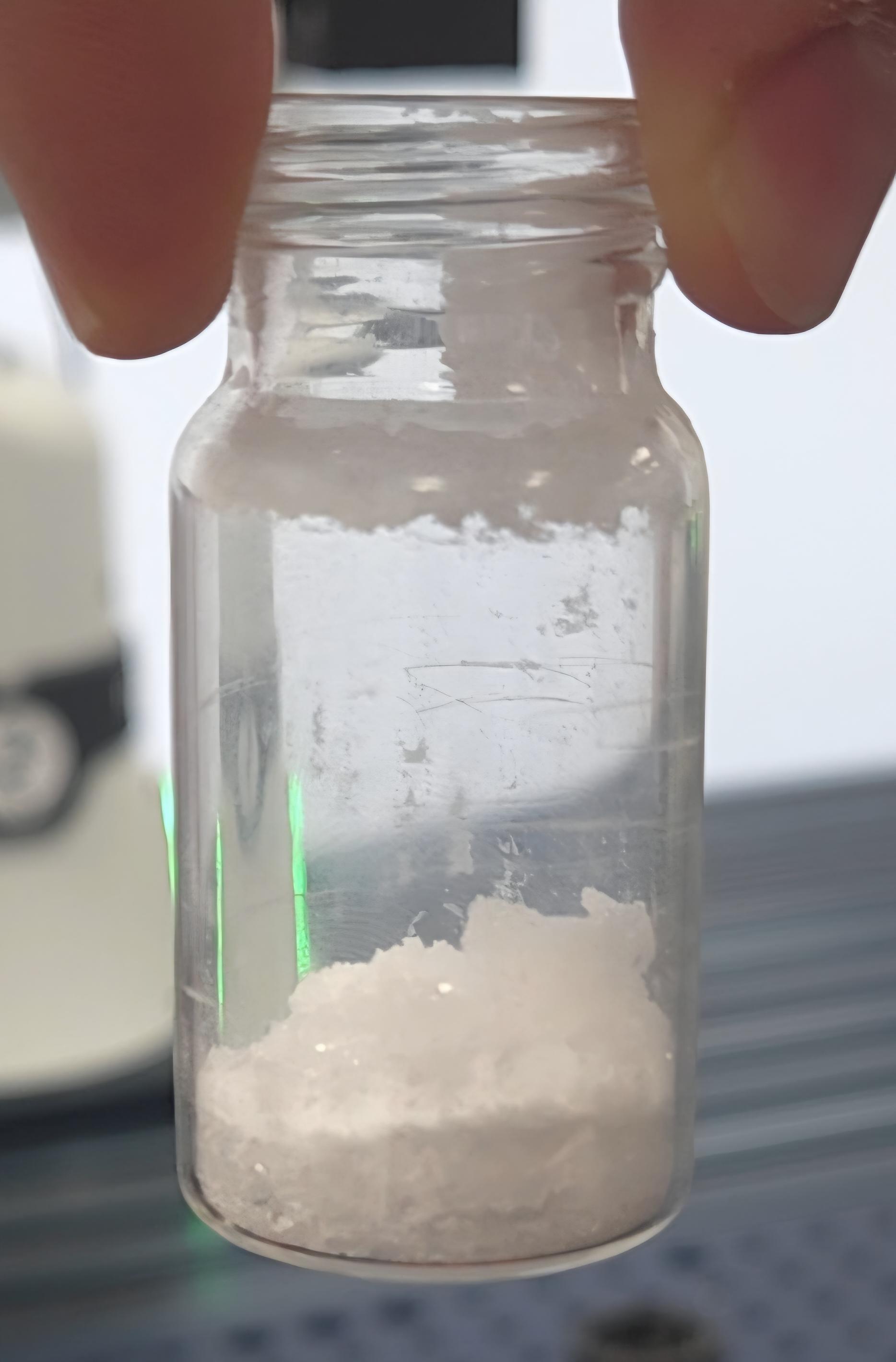} & \includegraphics[width=0.20\textwidth, height=2.5cm, keepaspectratio]{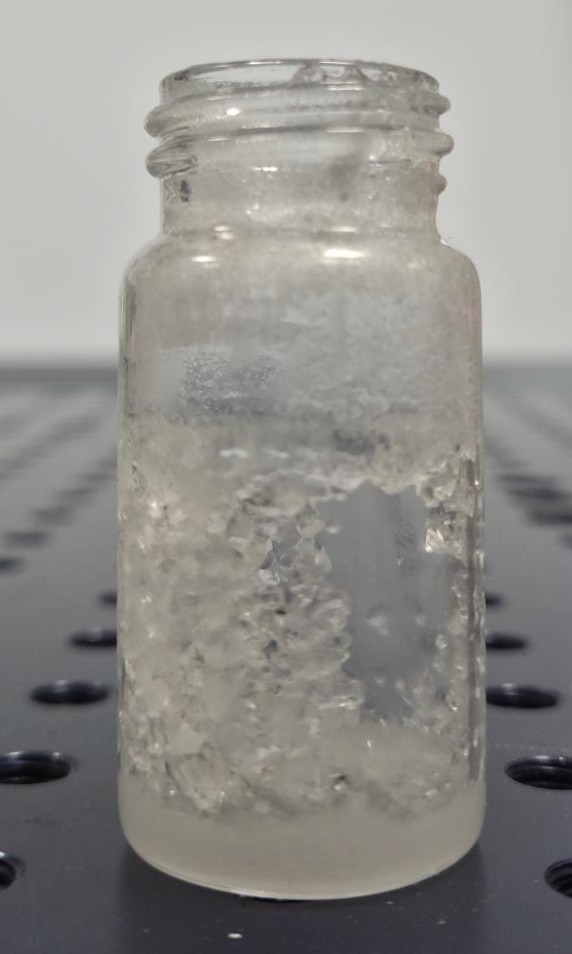} \\
        \bottomrule
    \end{tabular}
    \label{table:qual_results}
\end{table*}

\subsection{Robot Simulation Setup}
\label{ssec:robot_simulation}

\begin{figure}[h!] 
    \centering
    \begin{subfigure}[b]{0.46\columnwidth} 
        \centering
        \includegraphics[width=\textwidth]{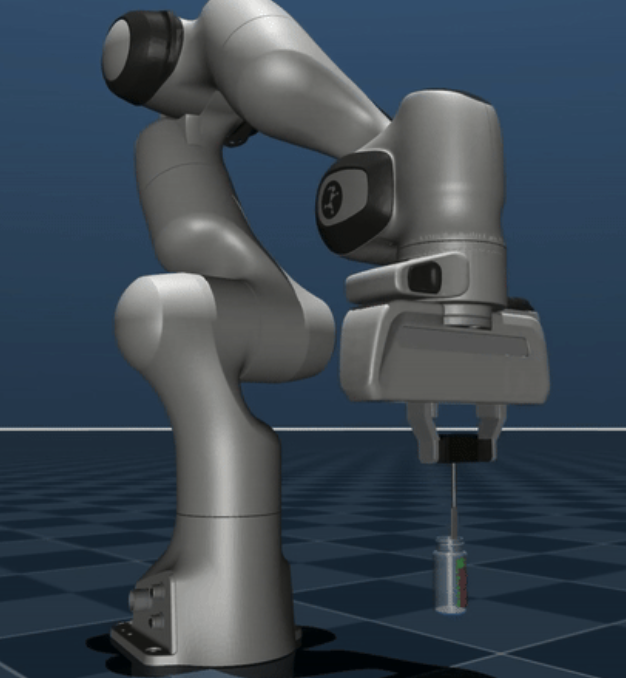} 
        \caption{Simulation environment.} 
        \label{fig:robot_sim}
    \end{subfigure}
    \hspace{0.5cm}
    \begin{subfigure}[b]{0.44\columnwidth} 
        \centering
        \includegraphics[width=\textwidth]{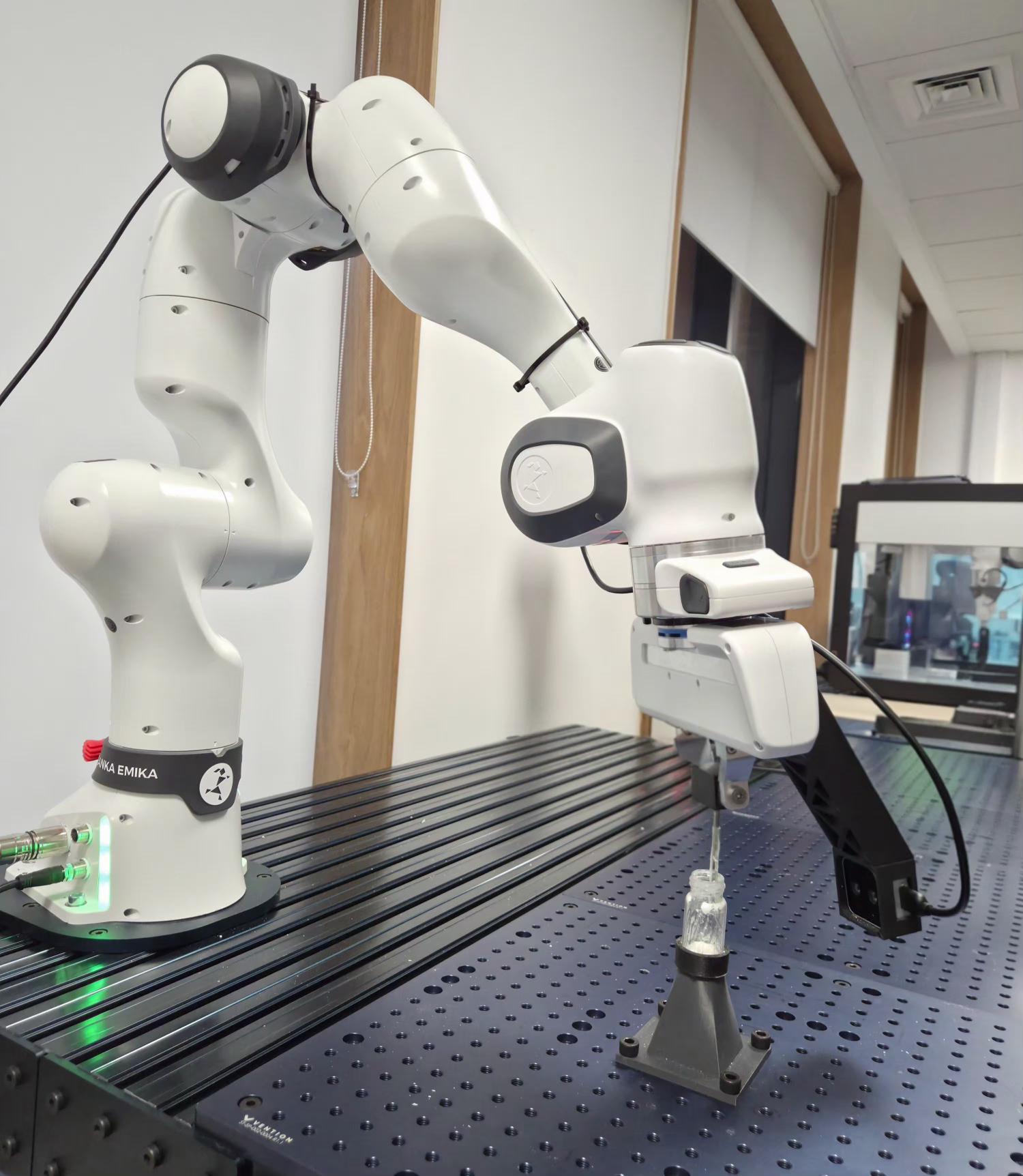} 
        \caption{Real robotic setup.}
        \label{fig:robot_real}
    \end{subfigure}
    \caption{An overview of the simulation environment and real robot setup, including the robotic manipulator (Franka Research 3), the scraping tool and the vial.}
    \label{fig:robot_setup}
\end{figure}

\begin{figure}[htbp!]
    \centering
    \includegraphics[width=0.48\textwidth]{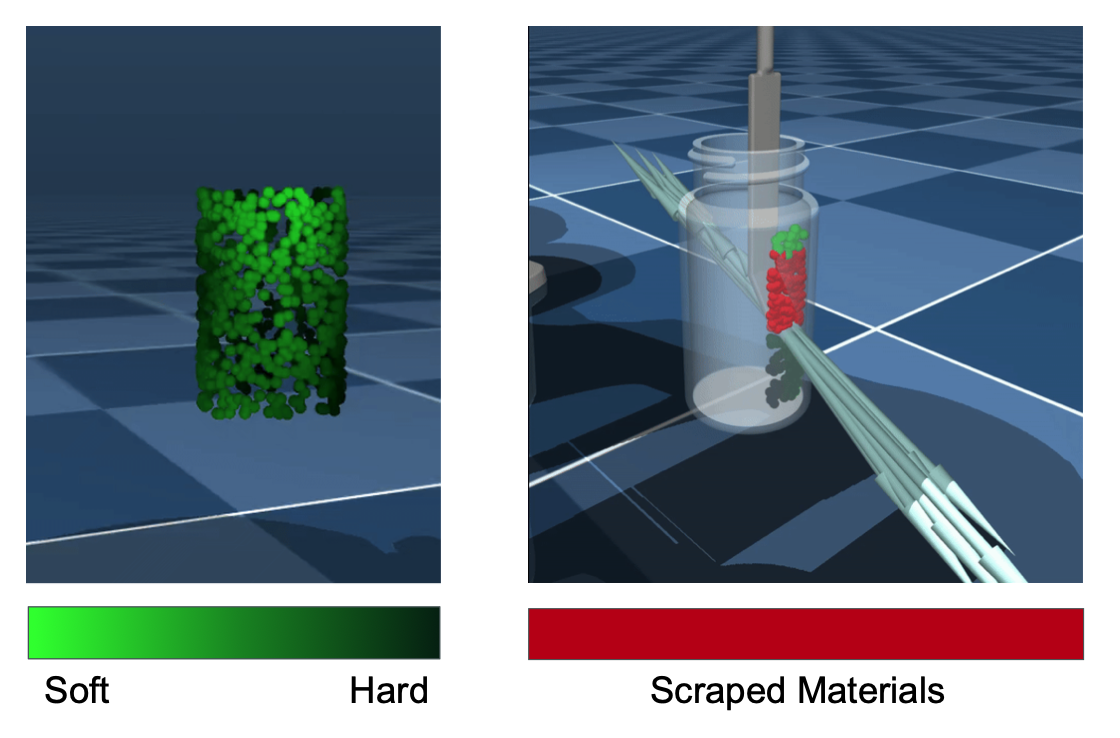}
    \caption{An overview of the simulated heterogeneous materials. Left: Particles are randomly generated with varying hardness levels, represented by shades of green (lighter green indicates softer, easier-to-dislodge particles; darker green indicates harder particles requiring greater force). Right: After scraping, successfully dislodged particles turn red, while remaining green particles represent material yet to be removed.}
    \label{fig:sim_env_spheres}
\end{figure}

Our simulated environment (Fig.~\ref{fig:robot_sim}) was developed using the MuJoCo physics simulator~\cite{Todorov2012}. 
It features a Franka Research 3 robot~\cite{Haddadin2022}, a scraping tool, and a sample vial containing heterogeneous materials. 
To facilitate the learning of an adaptive policy, we developed a novel particle-based model to represent these materials. 
Instead of a single entity, the sample is modelled as a collection of several hundred discrete spheres. 
To simulate diverse, unknown material characteristics, each sphere is assigned a unique dislodgement force threshold. 
These thresholds are procedurally generated using Perlin noise~\cite{perlin1985}, creating a rich and challenging training distribution. 
This method (Fig.~\ref{fig:sim_env_spheres}) allows the agent to learn a generalised scraping strategy that is not overfitted to a single material, which contributes to better sim-to-real transfer. 
% An illustration of this model is shown in Fig.~\ref{fig:sim_env_spheres}.
For the task, the robot began with the scraping tool already grasped to focus on the challenges of contact-rich insertion and scraping rather than grasping. 
The milestone rewards in Equation~\ref{eq:reward_function} were set to reward the agent when it reaches 50\% and 90\% task completion. 
In simulation, a task episode is deemed successful when the robot removes all material from the target window. 
The scraping tool, with dimensions matching the physical object, was accurately modelled within the simulation framework. 
The vial was represented as a solid object using a signed distance function (SDF)-based model. 
The simulation operates with a physics timestep of 0.001 s (1 kHz), while the Cartesian impedance controller receives commands at a control frequency of 500 Hz. 
This setup ensures a stable and accurate simulation of the robot's dynamics and its interaction with the environment. 
All simulation experiments were conducted on a machine equipped with an AMD Ryzen 9 9950X 32-Core CPU. 
We trained our policy exclusively in simulation and then transferred the model to the real robot using a zero-shot approach.

\subsection{Real Robotic Setup}
\label{ssec:real_robot_setup}

The real robotic setup, shown in Fig.~\ref{fig:robot_real}, consists of a Franka Research 3 robot~\cite{Haddadin2022} equipped with a Franka Hand parallel gripper. 
A 20 mL glass vial is placed in a 3D-printed holder, which serves a dual purpose: it elevates the vial to a height suitable for scraping while also ensuring its contents remain within the camera's field of view. 
For our task, a stainless steel spatula with a rounded tip was used as the scraping tool.

\subsection{Materials}
\label{ssec:materials}

To validate our system with a real robot, we used a series of different materials chosen for their challenging physical properties. 
The materials (Table~\ref{table:qual_results}) tested included evaporated crystalline solids (table salt and sugar), thick viscous pastes (made from flour or cornflour and water), and the resulting dry, hardened paste once the water had evaporated from the cornflour mixture.
For consistency, all materials were prepared in 20 mL vials and either left to dry naturally or heated on an IKA Plate (RCT digital) to accelerate evaporation of water. 
A key experimental challenge was the manual preparation of samples, which, combined with real-world environmental factors such as ambient temperature and humidity, led to variations in material properties and drying kinetics, even within the same type of substance.
To account for these variations, we introduce a \emph{relative success} metric, where $S_{\text{rel}} = {S_{\text{robot}}}/{S_{\text{human}}} \times 100\%$.
Here, $S_{\text{robot}}$ and $S_{\text{human}}$ are the gravimetrically measured scraping success rates for the robot and a human scientist, respectively. 
Given that absolute (100\%) material recovery is frequently precluded by vial geometry and surface adhesion, this human baseline establishes an empirical upper bound for performance evaluation.

% On the real system, achieving complete (100\%) removal is often infeasible due to vial geometry and material adhesion; we therefore adopt a relative success metric (Section~\ref{ssec:materials}) to evaluate real-world performance.

\subsection{Experiment I: Cartesian Impedance Controller (Baseline)}
\label{ssec:exp1_CIC}

We evaluate the baseline performance of a Cartesian impedance controller with a fixed wrench profile of 4~N across all materials(Section~\ref{ssec:materials}). 
This value was empirically determined from the mean interaction wrench of the converged RL policy (Section~\ref{ssec:exp3_RL}), ensuring the baseline represents an optimised fixed strategy.
Success was quantified gravimetrically by weighing vials before and after scraping. Each material was tested five times, with a human scientist also scraping after the robot for comparison.
Results are presented in Table~\ref{tab:experimental_results_CIC} and Table~\ref{table:qual_results}. 
The controller performed best with dried cornflour but struggled with crystalline materials, particularly sugar.
Since even manual scraping rarely achieved 100\% due to vial geometry and material adhesion, we report a relative success rate (Section~\ref{ssec:materials}), achieving 64.44\% overall.

\begin{table}[h!]
    \centering
    \caption{Results for fixed wrench profile (baseline method). Success rates averaged over 5 trials. }
    \label{tab:experimental_results_CIC}
    % Changed width back to 0.48\textwidth, which now fits
    \begin{tabular*}{0.49\textwidth}{@{\extracolsep{\fill}}lccc} 
        \toprule
        % Used \makecell to create multi-line headers
        Material & \makecell{Robot \\ Success (\%)} & \makecell{Human \\ Success (\%)} & \makecell{Relative \\ Success (\%)} \\
        \midrule
        Liquid Dough & $56.8 \pm 6.8$ & $90.1 \pm 4.5$ & $63.0 \pm 6.5$ \\
        Liquid Cornflour & $63.4 \pm 9.9$ & $92.5 \pm 2.9$ & $68.4 \pm 9.2$ \\
        Dried Cornflour & $83.3 \pm 8.9$ & $94.1 \pm 1.9$ & $88.6 \pm 9.6$ \\
        Crystalline Salt & $59.0 \pm 13.6$ & $96.2 \pm 4.2$ & $61.2 \pm 14.3$ \\
        Crystalline Sugar & $7.5 \pm 2.7$ & $19.1 \pm 8.5$ & $41.0 \pm 8.4$ \\
%        Crystalline Salt (1A) & $57.7 \pm 14.5$ & $97.6 \pm 1.8$ & $59.0 \pm 14.6$ \\
%        Crystalline Salt (2A) & $60.4 \pm 13.2$ & $94.9 \pm 5.2$ & $63.3 \pm 11.8$ \\
        \bottomrule
        \multicolumn{4}{l}{\footnotesize}
    \end{tabular*}
\end{table}

\subsection{Experiment II: Perception for Visual Monitoring of Materials in Sample Vials}
\label{ssec:exp2_perception}

To provide the RL agent with real-time visual feedback, the robot's end-effector was equipped with an Intel Realsense D405 RGB-D camera. 
This camera was selected for its high-resolution, short-range depth sensing and was mounted on a 3D-printed bracket 9 cm from the TCP to provide a clear, focused view of the vial's interior during scraping (Fig.~\ref{fig:robot_real}).

Given the camera's mounting on the robot arm, the vial's position relative to the camera was not constant, necessitating a robust segmentation method. 
The RGB image was first fed into a YOLOv8 model, pre-trained on the COCO dataset~\cite{Lin2014} and fine-tuned on 1,084 custom-labelled images of sample vials. 
If multiple vials were present, the centrally located vial was selected, as this was assumed to be the robot's focus. 
Following this, the GrabCut segmentation method was applied via the OpenCV library~\cite{Bradski2000}. Due to inconsistent depth readings from transparent glassware and crystals, artefacts with incorrect depth values were common. 
To correct these, we applied the smoothing method described in Equation~\ref{eqn:noise_correction}. 
Following depth thresholding (Fig.~\ref{fig:depth_thresholding}), we used OpenCV's contour finding to obtain a binary map of the material regions.
As mentioned in Section~\ref{ssec:perception}, the pipeline employs two stages of K-means clustering. 
The first stage uses six groups to robustly segment the spatula, providing sufficient colour granularity to distinguish its true hue from optically altered shades when viewed through translucent crystals. 
The spatula's green colour was chosen to create a large contrast with the materials in the HSV colour space, simplifying the perception pipeline. 
Specifically, any pixels within the mask whose hue fell within a 20-degree range of green (60 degrees in HSV colour space) were filtered out. 
After filtering the spatula, a second clustering stage uses three groups to produce a concise, static-sized representation of the material's distribution, which serves as a structured input for the RL policy.

In this evaluation, we defined a smaller viewing window by removing the top 25\% of the YOLO bounding box to exclude the threaded region, and 30\% from either end to focus on the central 40\% of the vial. 
We report the pipeline's performance with this smaller window on the most visually challenging material (crystalline) in Table~\ref{tab:perception_results}. 
While performance decreased by approximately 15\% after spatula removal, this trade-off is acceptable given the objective: to provide the robot with a high-precision approximation of the material's location rather than an accurate classification of all crystals. 
The model's high precision implies that the output, representing the location of remaining material, is a reliable and crucial input for the RL agent, informing its subsequent scraping actions.

\begin{table}[htbp]
\centering
\caption{Results for the perception method, averaged over 20 trials for crystalline materials, were evaluated using accuracy, precision, recall, specificity, and F1-score. The methods are: (1) the vial with the spatula (Spatula), (2) the vial without the spatula (No Spatula), and (3) the vial with the spatula algorithmically removed from the image (Filtered Spatula).}
\label{tab:perception_results}
\begin{tabular}{l c c c c c}
\hline
\textbf{Method} & \textbf{Acc.} & \textbf{Prec.} & \textbf{Rec.} & \textbf{Spec.} & \textbf{F1} \\
\hline
Spatula & 79.91\% & 78.94\%  & 82.91\%  & 76.75\%  & 80.88\%  \\
No Spatula & 77.19\%  & 74.09\%  & 86.71\%  & 66.74\%  & 79.91\%  \\
Filtered Spatula & 64.71\%  & 79.40\%  & 40.04\%  & 89.55\%  & 53.24\%  \\
\hline
\end{tabular}
\end{table}

%- Benchmark the baseline CIC, which uses predefined "Soft," "Medium," and "Hard" wrench profiles.

\subsection{Experiment III: Reinforcement Learning for Adaptive Wrench Generation Method}
\label{ssec:exp3_RL}

This experiment investigated whether an RL agent could learn to generate and adapt optimal interaction wrenches for sample scraping, particularly on materials with unknown physical properties. 
The agent's policy, trained using proximal policy optimisation (PPO)~\cite{schulman2017} within the Stable Baselines3 toolbox~\cite{Raffin2021}, was designed to output the optimal Cartesian wrench to maximise a cumulative reward for task success. 
The reward function in Equation~\ref{eq:reward_function} uses a material-removal-to-force efficiency term as the primary objective, augmented with a milestone bonus for near-complete scraping and a penalty for unintended collisions. 
This design encouraged the policy to apply the minimum force required for sustained progress while maintaining task completion and safe contact behaviour. 
Crucially, the efficiency-based reward was also instrumental in preventing a sim-to-real transfer failure mode where an explicit force penalty induced an oscillatory \textit{push-and-pull} behaviour that failed on the physical robot due to unmodelled static friction. 
The final policy successfully learned to apply the minimal necessary force while maintaining continuous contact. 
% , 
While the baseline method (Section~\ref{ssec:exp1_CIC}) achieved an average relative success rate of 64.44\%, our proposed method of using RL for adaptive wrench generation achieved an average relative success rate of 75.3\% across all materials, demonstrating its ability to adapt to varying physical properties.
% \begin{figure}[]
%     \centering
%     \includegraphics[width=0.49\textwidth]{figures/mean_force_profile.pdf}
%     \caption{Learned Force Profile of the RL Agent.}
%     \label{fig:mean_force_profile}
% \end{figure}

\begin{table}[h!]
    \centering
    \caption{Results for our method (RL for adaptive wrench generation). Success rates averaged over 5 trials. }
    \label{tab:experimental_results_RL}
    % Changed width back to 0.48\textwidth, which now fits
    \begin{tabular*}{0.49\textwidth}{@{\extracolsep{\fill}}lccc} 
        \toprule
        % Used \makecell to create multi-line headers
        Material & \makecell{Robot \\ Success (\%)} & \makecell{Human \\ Success (\%)} & \makecell{Relative \\ Success (\%)} \\
        \midrule
        Liquid Dough & $66.4 \pm 9.4$ & $88.5 \pm 5.5$ & $74.9 \pm 8.7$ \\
        Liquid Cornflour & $57.2 \pm 11.1$ & $82.1 \pm 13.3$ & $70.0 \pm 10.3$ \\
        Dried Cornflour & $84.2 \pm 2.6$ & $89.8 \pm 2.2$ & $93.8 \pm 1.8$ \\
        Crystalline Salt & $62.0 \pm 18.1$ & $85.7 \pm 16.2$ & $71.4 \pm 10.3$ \\
        Crystalline Sugar & $19.9 \pm 12.6$ & $28.5 \pm 15.0$ & $66.4 \pm 11.2$ \\
%        Crystalline Salt (1A) & $57.7 \pm 14.5$ & $97.6 \pm 1.8$ & $59.0 \pm 14.6$ \\
%        Crystalline Salt (2A) & $60.4 \pm 13.2$ & $94.9 \pm 5.2$ & $63.3 \pm 11.8$ \\
        \bottomrule
        \multicolumn{4}{l}{\footnotesize}
    \end{tabular*}
\end{table}

\subsection{Discussion}
\label{ssec:discussion}

The high inter-sample variance reflects real material heterogeneity rather than measurement noise, making fixed-force control unreliable: a setting tuned for one sample will undoubtedly underperform in denser regions.
Accordingly, learning interaction forces with RL was essential for handling complex dynamics, with the clearest gains on non-Newtonian material (\textit{e.g.}, liquid cornflour), where adaptive modulation reduced shear thickening and tool slippage; however, all methods remained limited on highly viscous and adhesive material (\textit{e.g.}, liquid dough).
Regarding potential tool fouling from accumulated material, the policy's state representation, which comprises visual feedback sphere cluster centroids and residue percentages, external wrench ($F_x, F_y, F_z, T_x, T_y, T_z$), Cartesian pose ($x, y, z, \phi, \theta, \psi$), and velocity ($\dot{x}, \dot{y}, \dot{z}$), provides sufficient information for the agent to adapt to any changes in tool dynamics. 
Furthermore, given the spatula's small mass (16 g) and the moderate applied forces (averaging 4 N), any material accumulation on the tool has negligible impact on the overall interaction dynamics.
% Potential tool fouling had limited impact in our setup because the policy state (visual cluster centroids/coverage, external wrench, Cartesian pose, and velocity) provides sufficient feedback to compensate for tool-dynamics variation, and because the spatula mass ($\sim16$~g) is small relative to the applied forces (about 4~N).
Results were strongest on crystalline samples, likely due to good correspondence between the discrete simulation model and rigid material behaviour, most notably for crystalline sugar where robot performance approached human performance.
The successful zero-shot sim-to-real transfer can be attributed to our comprehensive domain randomisation strategy. 
First, joint friction randomisation during training exposed the agent to diverse robot compliance characteristics, making it robust to differences between simulated and real manipulator dynamics. 
Second, randomising sphere hardness thresholds via Perlin noise implicitly modelled the variability in material adhesion and dislodgement forces. 
Third, procedurally randomising the spatial distribution of material prevented policy overfitting. 
Finally, the hierarchical control architecture, where a low-dimensional action space ($f_x$, $\tau_y$, $z^D$) commands a stable Cartesian impedance controller, isolates the RL agent from high-frequency dynamics that are difficult to simulate accurately.
% Finally, zero-shot sim-to-real transfer was supported by joint-friction randomisation, Perlin-noise sphere-hardness randomisation, procedural spatial randomisation, and a low-dimensional action space ($f_x^c$, $\tau_y^c$, $z^D$) executed by a stable Cartesian impedance controller.

% Additional videos of our material scraping experiments are available at: \url{https://anonymous.4open.science/w/learning_based_scraping-63C3}.

% \subsection{Experiment IV: Real-World Deployment}
% \label{ssec:exp4_RW_deployment}
% In this experiment, we study how  ...

% a policy trained to be both effective and force-efficient in simulation be successfully transferred to a physical robot, maintaining both its performance and its low-impact(gentle-contact) characteristics?

% The policy trained in the MuJoCo simulation will be deployed on the physical Franka Research 3 robot to perform scraping on 20 mL vial with real materials.

% \begin{table}[h!]
%     \centering
%     \caption{Experimental Results}
%     \label{tab:scraping_results_quant}
%     \begin{tabular}{lccccc}
%         \toprule
%         Method & Material 1 & Material 2 & Material 3 & Material 4 & Material 5 \\
%         \midrule
%         Manual Scraping Success Rate & $A \pm B$ & $A \pm B$ & $A \pm B$ & $A \pm B$ & $A \pm B$ \\
%         Robotic Scraping Success Rate (CIC) & $A \pm B$ & $A \pm B$ & $A \pm B$ & $A \pm B$ & $A \pm B$ \\
%         Robotic Scraping Success Rate (RL-CIC) & $A \pm B$ & $A \pm B$ & $A \pm B$ & $A \pm B$ & $A \pm B$ \\
%         \bottomrule
%     \end{tabular}
%     \label{table:quant_results}
% \end{table}

%% file: sections/conclusions.tex
\section{Conclusion}
\label{sec:conclusion} 

This paper presented a novel framework for force-aware sample scraping in chemistry labs, which integrates a Cartesian impedance controller for compliant physical interaction with an RL agent that intelligently generates appropriate interaction wrenches. 
By leveraging a perception-augmented state, our agent demonstrated adaptive behaviour and achieved robust sim-to-real transfer. 
We showed that our method successfully scrapes materials with diverse and previously unknown properties, a task unattainable by related work using fixed-wrench approaches. 
This successful integration of learned adaptive control with compliant execution is a significant step toward improving force-sensitive manipulation in robotic chemists, enhancing adaptability, and ultimately accelerating automated scientific experimentation in real-world lab environments.
% Future work will explore how ...
Future work will focus on several key areas to advance the system's capabilities, including enhancing the simulation of complex materials, such as slurries, and expand the agent's learned behaviours to handle a broader range of material dynamics. 
We will also investigate different spatula geometries to improve traction on challenging, low-friction crystalline structures and refine the simulation's physics to more accurately model real-world interactions.